\documentclass[letterpaper,10pt,conference]{ieeeconf}

\IEEEoverridecommandlockouts

\usepackage[utf8]{inputenc} 
\usepackage[T1]{fontenc}    
\usepackage{hyperref}       
\usepackage{url}            
\usepackage{booktabs}       
\usepackage{amsfonts}       
\usepackage{nicefrac}       
\usepackage{microtype}      

\usepackage{amsmath}
\usepackage{amssymb}

\usepackage{dblfloatfix}
\usepackage{graphicx} 
\usepackage[font=footnotesize]{caption}
\usepackage{subcaption}
\usepackage{wrapfig}
\usepackage{color}

\usepackage{algorithm}
\usepackage{algorithmic}

\usepackage[export]{adjustbox}

\newcommand{\bs}{\mathbf{s}}
\newcommand{\ba}{\mathbf{a}}
\newcommand{\dt}{{\Delta}t}

\newcommand{\benv}{\mathbf{e}}
\newcommand{\bim}{\mathbf{I}}

\usepackage{gensymb}

\newcommand{\specialcell}[2][c]{%
  \begin{tabular}[#1]{@{}c@{}}#2\end{tabular}}

\title{\LARGE \bf
Learning Image-Conditioned Dynamics Models\\for Control of Under-actuated Legged Millirobots
}

\author{
    Anusha Nagabandi, Guangzhao Yang, Thomas Asmar, Ravi Pandya,\\%
    Gregory Kahn, Sergey Levine, Ronald S. Fearing\\
    University of California, Berkeley
}

\begin{document}

\maketitle
\thispagestyle{empty}
\pagestyle{empty}



\begin{abstract}
Millirobots are a promising robotic platform for many applications due to their small size and low manufacturing costs. Legged millirobots, in particular, can provide increased mobility in complex environments and improved scaling of obstacles. However, controlling these small, highly dynamic, and underactuated legged systems is difficult. Hand-engineered controllers can sometimes control these legged millirobots, but they have difficulties with dynamic maneuvers and complex terrains. We present an approach for controlling a real-world legged millirobot that is based on learned neural network models. Using less than 17 minutes of data, our method can learn a predictive model of the robot's dynamics that can enable effective gaits to be synthesized on the fly for following user-specified waypoints on a given terrain. Furthermore, by leveraging expressive, high-capacity neural network models, our approach allows for these predictions to be directly conditioned on camera images, endowing the robot with the ability to predict how different terrains might affect its dynamics. This enables sample-efficient and effective learning for locomotion of a dynamic legged millirobot on various terrains, including gravel, turf, carpet, and styrofoam.
Experiment videos can be found at \url{https://sites.google.com/view/imageconddyn}
\end{abstract}


\section{Introduction}

Legged millirobots are an effective platform for applications, such as exploration, mapping, and search and rescue, because their small size and mobility allows them to navigate through complex, confined, and hard-to-reach environments that are often inaccessible to aerial vehicles and untraversable by wheeled robots. Millirobots also provide additional benefits in the form of low power consumption and low manufacturing costs, which enables scaling them to large teams that can accomplish more complex tasks. This superior mobility, accessibility, and scalability makes legged millirobots some of the most mission-capable small robots available. However, the same properties that enable these systems to traverse complex environments are precisely what make them difficult to control.

Modeling the hybrid dynamics of under-actuated legged millirobots from first principles is exceedingly difficult due to complicated ground contact physics that arise while moving dynamically on complex terrains. Furthermore, cheap and rapid manufacturing techniques cause each of these robots to exhibit varying dynamics. Due to these modeling challenges, many locomotion strategies for such systems are hand-engineered and heuristic. These manually designed controllers impose simplifying assumptions, which not only constrain the overall capabilities of these platforms, but also impose a heavy burden on the engineer. Additionally, and perhaps most importantly, they preclude the opportunity for adapting and improving over time.

\begin{figure}[t]
    \centering
    \includegraphics[height=0.2\textheight]{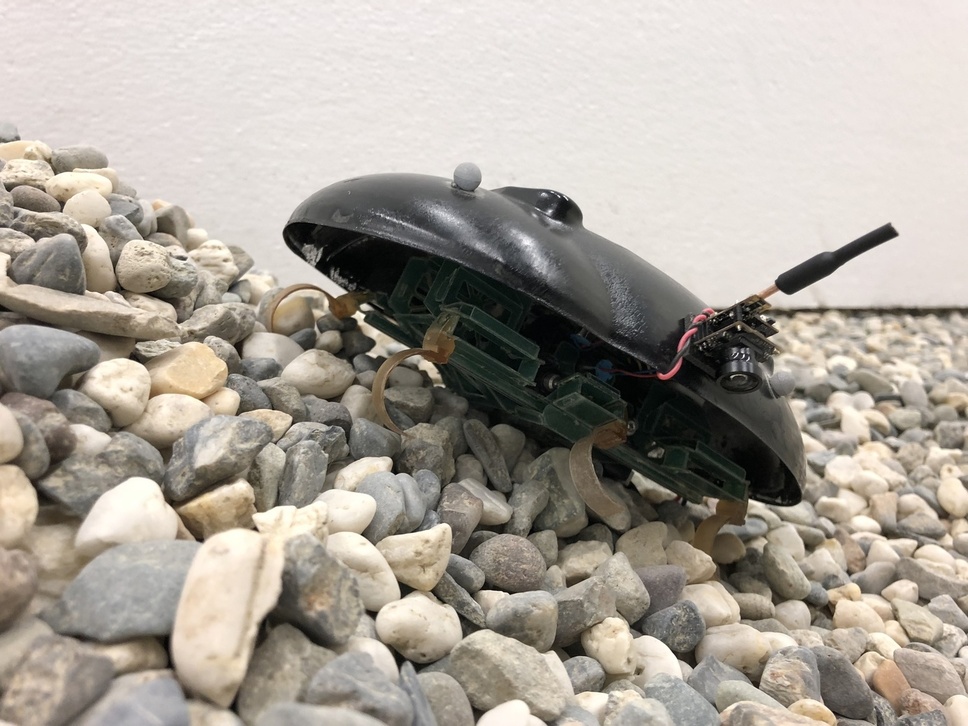}
    \caption{VelociRoACH: the small, mobile, highly dynamic, and bio-inspired hexapedal millirobot used in this work, shown with a camera mounted for terrain imaging.
    }
    \label{fig:roach}
    \vspace{-10pt}
\end{figure}

In this paper, we explore how learning can be used to automatically acquire locomotion strategies in diverse environments for small, low-cost, and highly dynamic legged robots. Choosing an appropriate learning algorithm requires consideration of a number of factors. First, the learned model needs to be expressive enough to cope with the highly dynamic and nonlinear nature of legged millirobots, as well as with high-dimensional sensory observations such as images. Second, the algorithm must allow the robot to learn quickly from modest amounts of data, so as to make it a practical algorithm for real-world application. Third, the learned general-purpose models must be able to be deployed on a wide range of navigational tasks in a diverse set of environments, with minimal human supervision.

The primary contribution of our work is an approach for controlling dynamic legged millirobots that learns an expressive and high-dimensional image-conditioned neural network dynamics model, which is then combined with a model predictive controller (MPC) to follow arbitrary paths. Our sample efficient learning-based approach uses less than 17 minutes of real-world data to learn to follow desired paths in a desired environment, and we empirically show that it outperforms a conventional differential drive control strategy for highly dynamic maneuvers. Our method also enables adaptation to diverse terrains by conditioning its dynamics predictions on its own observed images, allowing it to predict how terrain features such as gravel or turf will alter the system's response. To the best of our knowledge, we believe this work is the first to leverage and build upon recent advances in learning to achieve a high-performing and sample efficient approach for controlling dynamic legged millirobots.

\section{Related Work}
\label{sec:relatedwork}

\textbf{Controlling Legged Millirobots:} Extensive prior work on controlling legged robots includes larger legged robots such as {\em Anymal}~\cite{hutter2016anymal}, {\em ASIMO}~\cite{asimo_sakagami2002intelligent}, and {\em Big Dog}~\cite{raibert2008bigdog}. These systems can achieve successful locomotion, but they have multiple degrees of freedom per leg and a relatively slow stride frequency that allows for more sophisticated control strategies of planned foot placement~\cite{byl2008metastable,kolter2008hierarchical,kalakrishnan2010fast,zucker2011optimization}. Other prior work includes systems such as RHex~\cite{rhex}, where each leg has an independent actuator and can thus execute stable alternating tripod gaits to achieve desired motion. Unlike these systems, however, we are interested in dynamic legged millirobots that are underactuated; these descriptors imply that we cannot move each leg independently, that we have neither the ability nor time to plan specific foot placement, and that we cannot strive for static or quasi-static gaits where stability and well-behaved dynamics can be expected. This realm of steering methods for dynamic running of underactuated legged millirobots includes various methods~\cite{mcclung2006techniques,zarrouk2015dynamic}, such as actively changing leg kinematics~\cite{sprawlita,isprawl}, modulating leg impedance~\cite{dynaroach}, and executing roll oscillation modulated turning~\cite{haldane2014roll}. However, these approaches achieve open-loop turning gaits, while we desire a closed-loop approach to precise path execution. Other traditional methods for both control and modeling of legged systems make simplifying assumptions, such as approximating a system as a spring loaded inverted pendulum (SLIP) model~\cite{runningslip,hexapodslip} or approximating a system's behavior with a differential drive control strategy. Although these approaches do succeed in certain regimes~\cite{rhexslip}, they fail when high speeds or irregular environments lead to more complicated dynamics. In contrast, our neural network learning-based approach can cope with complex dynamics, while also incorporating high-dimensional environmental information in the form of images.

\textbf{Gait Optimization:} Instead of building on simplifying model assumptions to design controllers, prior work has also explored various methods of automatic gait optimization~\cite{Da2017,Gay2013}. These methods include stochastic gradient descent~\cite{tedrake2005learning}, genetic algorithms~\cite{chernova2004evolutionary}, and Bayesian optimization~\cite{calandra2014experimental,lizotte2007automatic,tesch2011using} to reduce the time-consuming design process of manually finding robust parameters. For instance,~\cite{tedrake2005learning} optimized a control policy for bipedal walking online in less than 20 minutes on a simplified system with 6 joints, and~\cite{Gay2013} learned model-free sensory feedback controllers to supplement specified open-loop gaits. While these methods are sample efficient and can be applied to real systems, they have not yet been shown to work for high dimensional systems or more complex systems, such as fast robots operating in highly dynamic regimes on irregular surfaces with challenging contact dynamics.

\textbf{Model-free Policy Learning:} Rather than optimizing gaits, prior work in model-free reinforcement learning algorithms has demonstrated the ability to instead learn these behaviors from scratch. Work in this area, including Q-learning~\cite{mnih2015human,Oh2016_ICML}, actor-critic methods~\cite{Lillicrap2016_ICLR,Mnih2016_ICML}, and policy gradients~\cite{Schulman2015_ICML}, has learned complex skills in high-dimensional state spaces, including skills for simulated robotic locomotion tasks. However, the high sample complexity of such purely model-free algorithms makes them difficult to use for learning in the real world, where sample collection is limited by time and other physical constraints. To our knowledge, no prior method has attempted model-free deep reinforcement learning of locomotion skills in the real-world, but Gu et al.~\cite{shane_gu_ethan_holly_icra_2017} learn reaching skills with a robotic arm using several hours of experience. Unlike these approaches, our model-based learning method uses only minutes of experience to achieve generalizable real-world locomotion skills that were not explicitly seen during training, and it further exemplifies the benefits in sample complexity that arise from incorporating models with learning-based approaches.

\textbf{Model Learning:} Although the sample efficiency of model-based learning is appealing, and although data-driven approaches can eliminate the need to impose restrictive assumptions or approximations, the challenge lies in the difficulty of learning a good model. Relatively simple function approximators such as time-varying linear models have been used to model dynamics of systems~\cite{lioutikov2014sample,yip2014model}, including our VelociRoACH~\cite{buchan2013automatic} platform. However, these models have not yet been shown to posses enough representational power (i.e., accuracy) to generalize to complex locomotion tasks. Prior work has also investigated learning probabilistic dynamics models~\cite{Deisenroth2011_ICML,ko2008gp}, including Gaussian process models for simulated legged robots~\cite{deisenroth2012toward}. However, to the best of our knowledge, no prior work has learned Gaussian process models for real-time control of dynamic real-world legged robots from raw data. Also, while these approaches can be sample efficient, it is intractable to scale them to higher dimensions, as needed especially when incorporating rich sensory inputs such as image observations. In contrast, our method employs expressive neural network dynamics models, which easily scale to high dimensional inputs. Other modeling approaches have leveraged smaller neural networks for dynamics modeling, but they impose strict and potentially restrictive structure to their formulation, such as designing separate modules to represent the various segments of a stride~\cite{Crusea1998}, approximating actuators as muscles and tuning these parameters~\cite{Xiong2014}, or calculating equations of motion and learning error terms on top of these specific models~\cite{Grandia2018}. Instead, we demonstrate a sample efficient, expressive, and high-dimensional neural network dynamics model that is free to learn without the imposition of an approximated hand-specified structure.

\textbf{Environment Adaptation:} The dynamics of a robot depend not only on its own configuration, but also on its environment. Prior methods generally categorize the problem of adapting to diverse terrains into two stages: first, the terrain is recognized by a classifier trained with human-specified labels (or, less often, using unsupervised learning methods~\cite{lefflerThesis}), and second, the gait is adapted to the terrain. This general approach has been used for autonomous vehicles~\cite{thrun2006stanley,lefflerThesis}, larger legged robots~\cite{kolter2008hierarchical,kalakrishnan2010fast,zucker2011optimization,Xiong2014,hoepflinger2010haptic}, and for legged millirobots~\cite{wu2016integrated, bermudez2012performance}. In contrast, our method does not require any human labels at run time, and it adapts to terrains based entirely on autonomous exploration: the dynamics model is simply conditioned on image observations of the terrain, and it automatically learns to recognize the visual cues of terrain features that affect the robot's dynamics.

\textbf{This work:} While our prior work evaluated model-based reinforcement learning with neural network models~\cite{nagabandi2017neural}, to our knowledge, the present work is the first to extend these model-based learning techniques to real-world robotic locomotion on various terrains. Furthermore, we present a novel extension of this approach that conditions the dynamics predictions on image observations and allows for adaptation to various terrain types.
\section{Model-Based Learning Method for Locomotion Control}
\label{sec:methods}

\begin{figure}[b]
    \centering
    \includegraphics[width=\columnwidth]{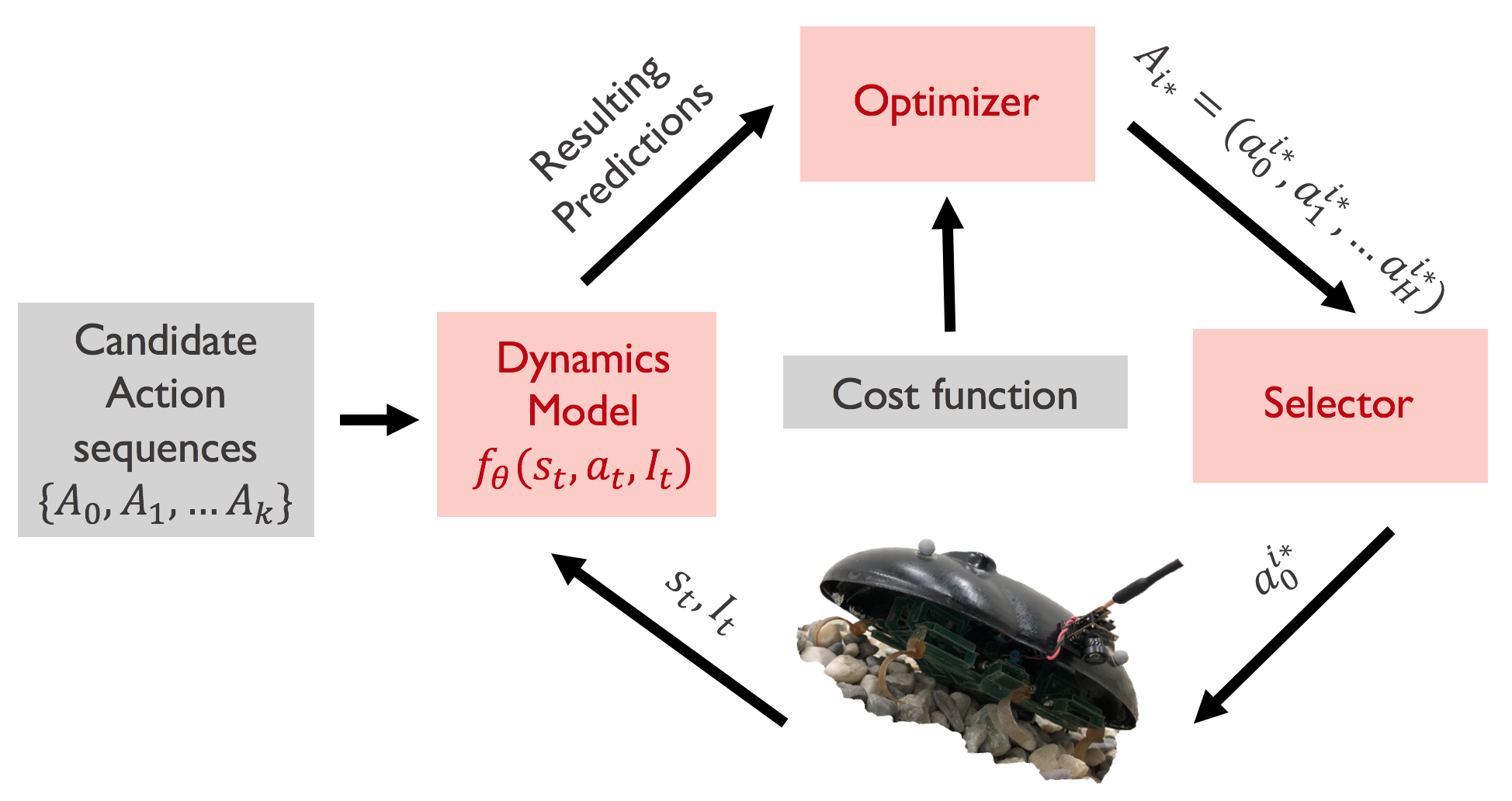}
    \caption{Our image-conditioned model-based learning method for locomotion control: A closed-loop MPC controller uses predictions from the learned dynamics model 
    to perform action selection.}
    \label{fig:blockdiagram}
\end{figure}

\begin{figure*}[t]
    \centering
    \includegraphics[width=\textwidth]{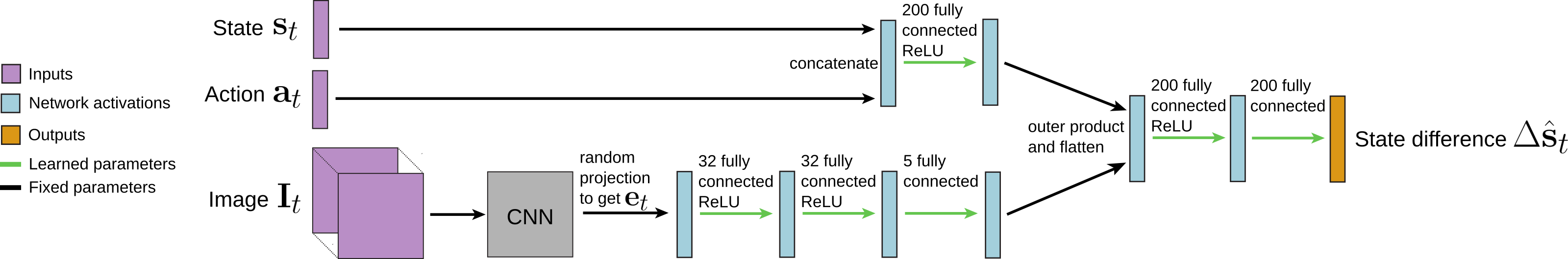}
    \caption{Our image-conditioned neural network dynamics model. The model takes as input the current state $\bs_t$, action $\ba_t$, and image $\bim_t$. The image is passed through the convolutional layers of AlexNet~\cite{krizhevsky2012imagenet} pre-trained on ImageNet~\cite{deng2009imagenet}, which is then flattened and projected into a lower dimension through multiplication with a random fixed matrix to obtain $\benv_t$. The image and concatenated state-action vectors are passed through fully connected layers, fused via an outer product, flattened, and passed through more fully connected layers to obtain the predicted state difference $\Delta \hat{\bs}_{t}$.}
    \label{fig:terrain-conditional}
\end{figure*}

In this work, we propose an automated method of acquiring locomotion strategies for small, low-cost, dynamic legged millirobots. 
In this section, we describe a method for learning a neural network dynamics model~\cite{nagabandi2017neural} (Sec.~\ref{sec:model}), using the model as part of a model predictive controller (Sec.~\ref{sec:controller}), and extending the model into an image-conditioned model using features from a pre-trained convolutional neural network. Fig.~\ref{fig:blockdiagram} provides an overview of our method.


\subsection{Learning System Dynamics}
\label{sec:model}

We require a parameterization of the dynamics model that can cope with high-dimensional state and action spaces, and the complex dynamics of legged millirobots. We therefore represent the dynamics function $\hat{f}_\theta(\bs_t, \ba_t)$ as a multilayer
neural network, parameterized by $\theta$. This function outputs the predicted change in state that occurs as a result of executing action $\ba_t$ from state $\bs_t$, over the time step duration of $\dt$. Thus, the predicted next state is given by: $\hat{\bs}_{t+1} = \bs_t + \hat{f}_\theta(\bs_t, \ba_t)$. While choosing too small of a $\dt$ leads to too small of a state difference to allow meaningful learning, increasing the $\dt$ too much can also make the learning process more difficult because it increases the complexity of the underlying continuous-time dynamics. Although we do not perform a structured study of various $\dt$ values for our system, we provide this insight as something for consideration when implementing this method on other systems.

We define the state $\bs_t$ of the VelociRoACH to be $[x,~y,~z,~v_x,~v_y,~v_z,~\cos(\phi_r),~\sin(\phi_r),~\cos(\phi_p),~\sin(\phi_p),$ $~\cos(\phi_y),~\sin(\phi_y),~\omega_x,~\omega_y,~\omega_z,~\cos(aL),~\sin(aL),$ $~\cos(aR),~\sin(aR),~v_{aL},~v_{aR},~{\rm bemf}_L,~{\rm bemf}_R,~V_{bat}]^T$. The center of mass positions $(x,y,z)$ and the Euler angles to describe the center of mass pose $(\phi_r, \phi_p, \phi_y)$ come from the OptiTrack motion capture system. The angular velocities $(\omega_x,\omega_y,\omega_z)$ come from the gyroscope onboard the IMU, and the motor crank positions $(aL,aR)$ come from the magnetic rotary encoders, which give a notion of leg position. We include $({\rm bemf}_L,{\rm bemf}_R)$ because back-EMF provides a notion of motor torque/velocity, and $(V_{bat})$ because the voltage of the battery affects the VelociRoACH's performance. Note that the state includes $\sin$ and $\cos$ of angular values, which is common practice and allows the neural network to avoid wrapping issues.


We define the action representation of the VelociRoACH to represent desired velocity setpoints for the rotation of the legs, and we achieve these setpoints using a lower-level PID controller onboard the system. We collect training data by placing the robot in arbitrary start states and executing random actions at each time step. We record each resulting trajectory ${\tau = (\bs_0, \ba_0, \cdots, \bs_{T-2}, \ba_{T-2}, \bs_{T-1})}$ of length $T$. We slice the trajectories $\{\tau\}$ into training data inputs $(\bs_t, \ba_t)$ and corresponding output labels $(\bs_{t+1} - \bs_t)$. We train the dynamics model $\hat{f}_\theta(\bs_t, \ba_t)$ on data from the training dataset $\mathcal{D}$ by minimizing the error
\begin{align}
\mathcal{E}(\theta) = \frac{1}{|\mathcal{D}|} \sum_{(\bs_t, \ba_t, \bs_{t+1}) \in \mathcal{D}} \frac{1}{2} \| (\bs_{t+1} - \bs_t) - \hat{f}_\theta(\bs_t, \ba_t) \|^2_2 \label{eqn:our-mse}
\end{align}
using stochastic gradient descent. Prior to training, we preprocess the training data by normalizing it to be mean $0$ and standard deviation $1$, which ensures equal weighting of different state elements, regardless of their magnitudes.


\subsection{Model-Based Control Using Learned Dynamics}
\label{sec:controller}

We formulate a model-based controller which uses the learned model $\hat{f}_\theta(\bs_t, \ba_t)$ together with a cost function $c(\bs_t, \ba_t)$ that encodes some task. Many methods could be used to perform this action selection, and we use a random-sampling shooting method~\cite{Rao2009_shooting}. At each time step $t$, we randomly generate $K$ candidate action sequences of $H$ actions each, use the learned dynamics model to predict the resulting states, and then use the cost function to select the action sequence with the lowest cost. The cost function that we use for path following is as follows:
\begin{align}
c(\bs_t, \ba_t) = f_p*p + f_h*h + f_f*f,
\label{eqn:cost}
\end{align}
where the parameter $f_p$ penalizes perpendicular distance $p$ away from the desired path, parameter $f_f$ encourages forward progress $f$ along the path, and parameter $f_h$ maintains the heading $h$ of the system toward the desired direction. Rather than executing the entire sequence of selected optimal actions, we use model predictive control (MPC) to execute only the first action $\ba_t$, and we then replan at the next time step, given updated state information. 


\subsection{Image-Conditioned Dynamics Model}
\label{sec:terrainCond}

As currently described, our model-based learning approach can successfully follow arbitrary paths when trained and tested on a single terrain. However, in order to traverse complex and varied terrains, it is necessary to adjust the dynamics to the current terrain conditions. One approach to succeeding in multiple environments would be to train a separate dynamics model for each terrain. However, in addition to requiring many separate models, this would lead to models that would likely generalize poorly. Furthermore, this approach would require a person to label the training data, as well as each run at test-time, with which terrain the robot is in. All of these aspects are undesirable for an autonomous learning system.

Instead, we propose a simple and highly effective method for incorporating terrain information, using only observations from a monocular color camera mounted on the robotic platform. We formulate an image-conditioned dynamics model $\hat{f}_\theta(\bs_t, \ba_t, \bim_t)$ that takes as input not only the current robot state $\bs_t$ and action $\ba_t$, but also the current image observation $\bim_t$. The model (Fig.~\ref{fig:terrain-conditional}) passes image $\bim_t$ through the first eight layers of AlexNet~\cite{krizhevsky2012imagenet}. The resulting activations are flattened into a vector, and this vector is then multiplied by a fixed random matrix in order to produce a lower dimensional feature vector $\benv_t$. The concatenated state-action vector $[\bs_t; \ba_t]$ is passed through a hidden layer and combined with $\benv_t$ through an outer product. As opposed to a straightforward concatenation of $[\bs_t; \ba_t; \benv_t]$, this outer product allows for higher-order integration of terrain information terms with the state and action information terms. This combined layer is then passed through another hidden layer and output layer to produce a prediction of state difference $\Delta \hat{\bs}_{t}$.

Training the entire image-conditioned neural network dynamics model with only minutes of data---corresponding to tens of thousands of datapoints---and in only a few environments would result in catastrophic overfitting. Thus, to perform feature extraction on our images, we use the AlexNet~\cite{krizhevsky2012imagenet} layer weights optimized from training on the task of image classification on the ImageNet~\cite{deng2009imagenet} dataset, which contains 15 million diverse images. Although gathering and labelling this large image dataset was a significant effort, we note that such image datasets are ubiquitous and their learned features have been shown to transfer well to other tasks~\cite{razavian2014cnn}. By using these pre-trained and transferable features, our image-conditioned dynamics model is sample-efficient and can automatically adapt to different terrains without any manual labelling of terrain information.

We show in our experiments that this image-conditioned dynamics model outperforms a na\"{i}vely trained dynamics model that is trained simply on an aggregation of all the data. Furthermore, the performance of our image-conditioned dynamics model is comparable, on each terrain, to individual dynamics models that are specifically trained (and tested) on that terrain.


\section{Results}
\label{sec:results}

\begin{figure}[!b]
    \centering
    \includegraphics[trim={0 1cm 0 1cm},clip,width=0.85\columnwidth]{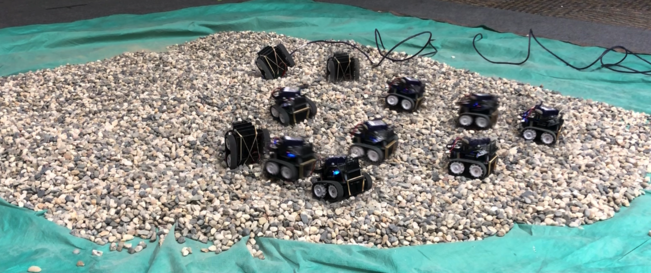}
    \includegraphics[trim={0 1cm 0 1cm},clip,width=0.85\columnwidth]{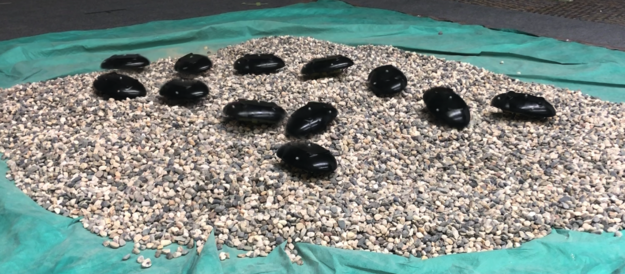}
    \caption{Over 15 teleoperated trials performed on rough terrain, a legged robot succeeded in navigating through the terrain 90\% of the time, whereas a wheeled robot of comparable size succeeded only 30\% of the time.}
    \label{fig:wheeled}
\end{figure}

\begin{figure*}[t]
    \centering
    \includegraphics[height=0.12\textheight]{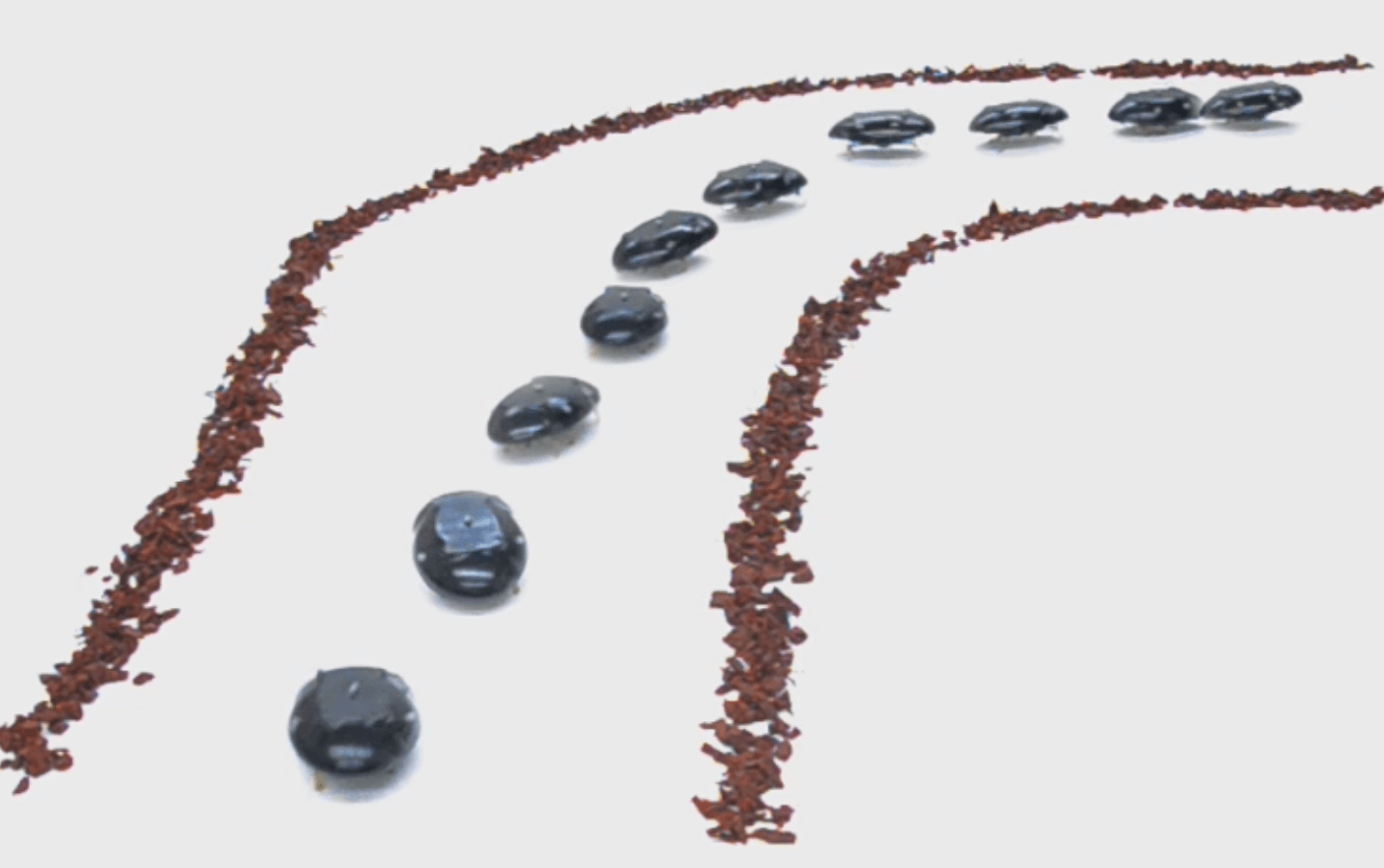}
    \includegraphics[height=0.12\textheight]{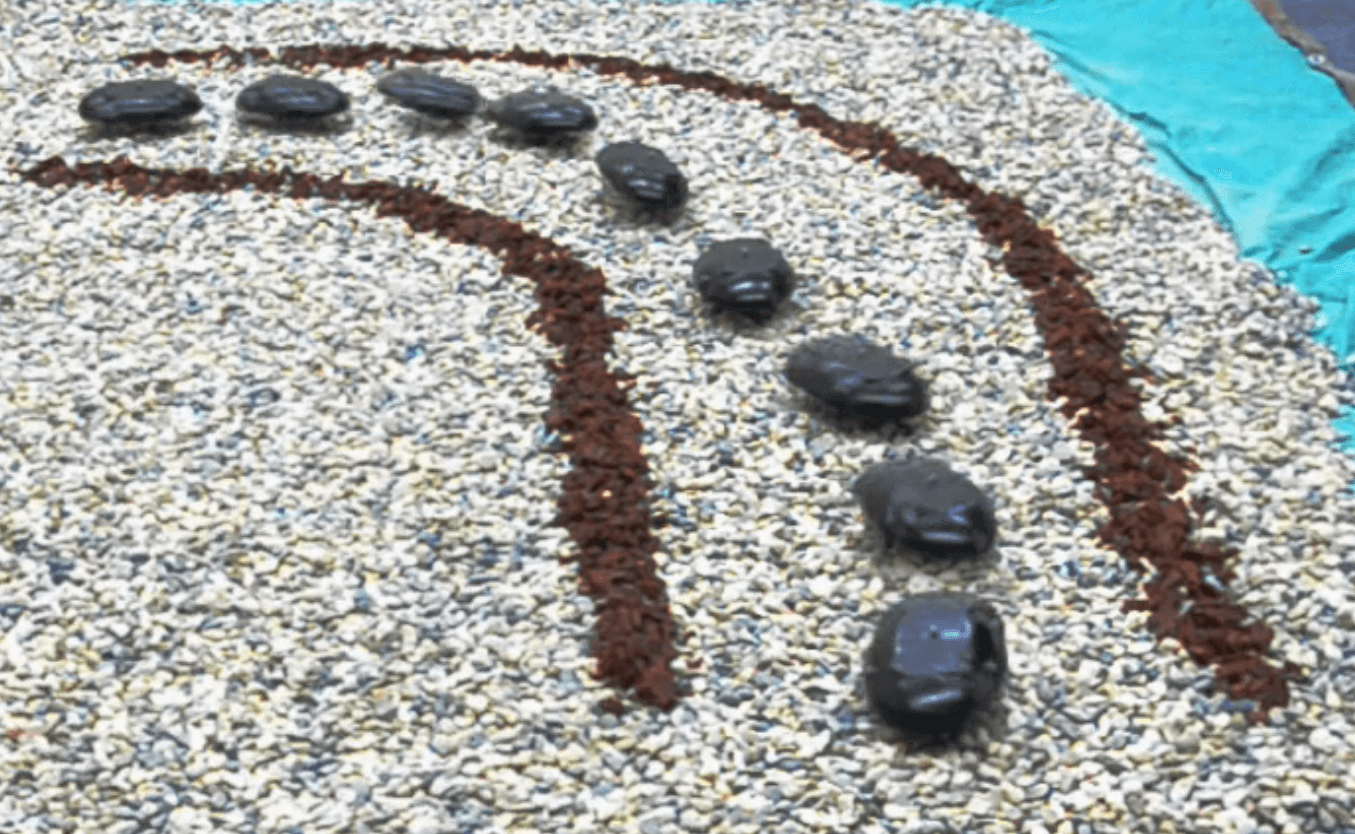}
    \includegraphics[height=0.12\textheight]{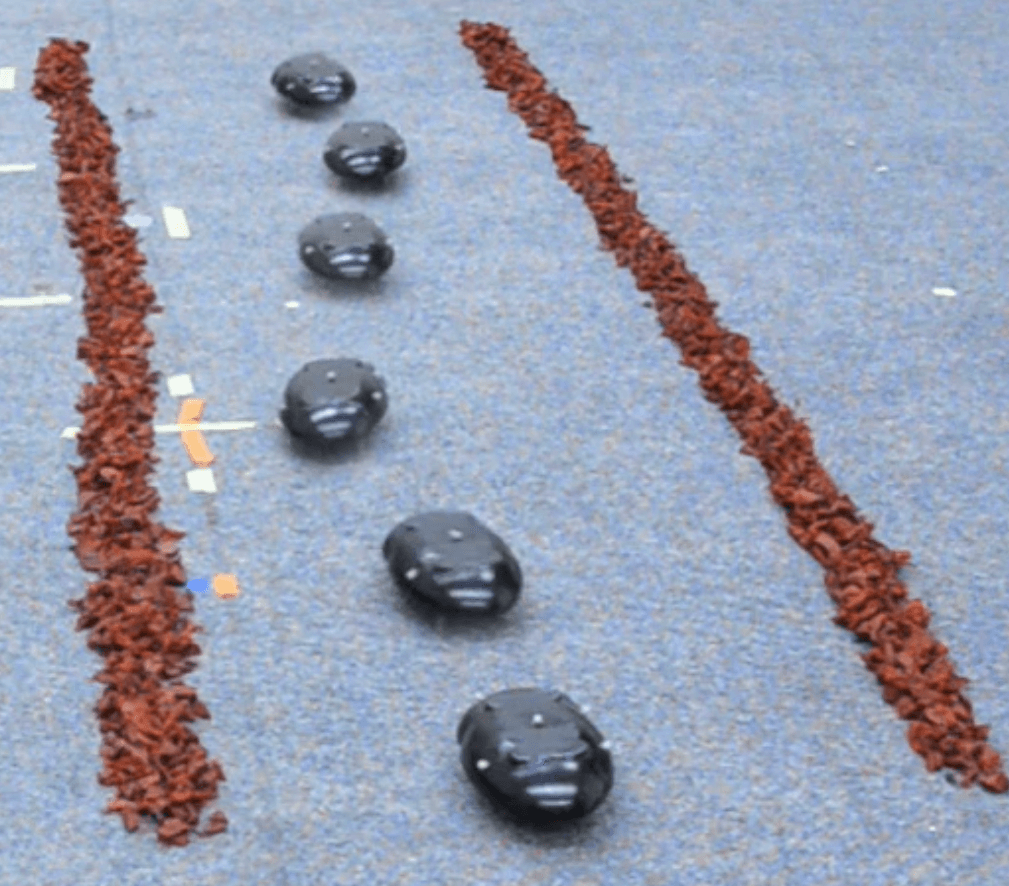}
    \includegraphics[height=0.12\textheight]{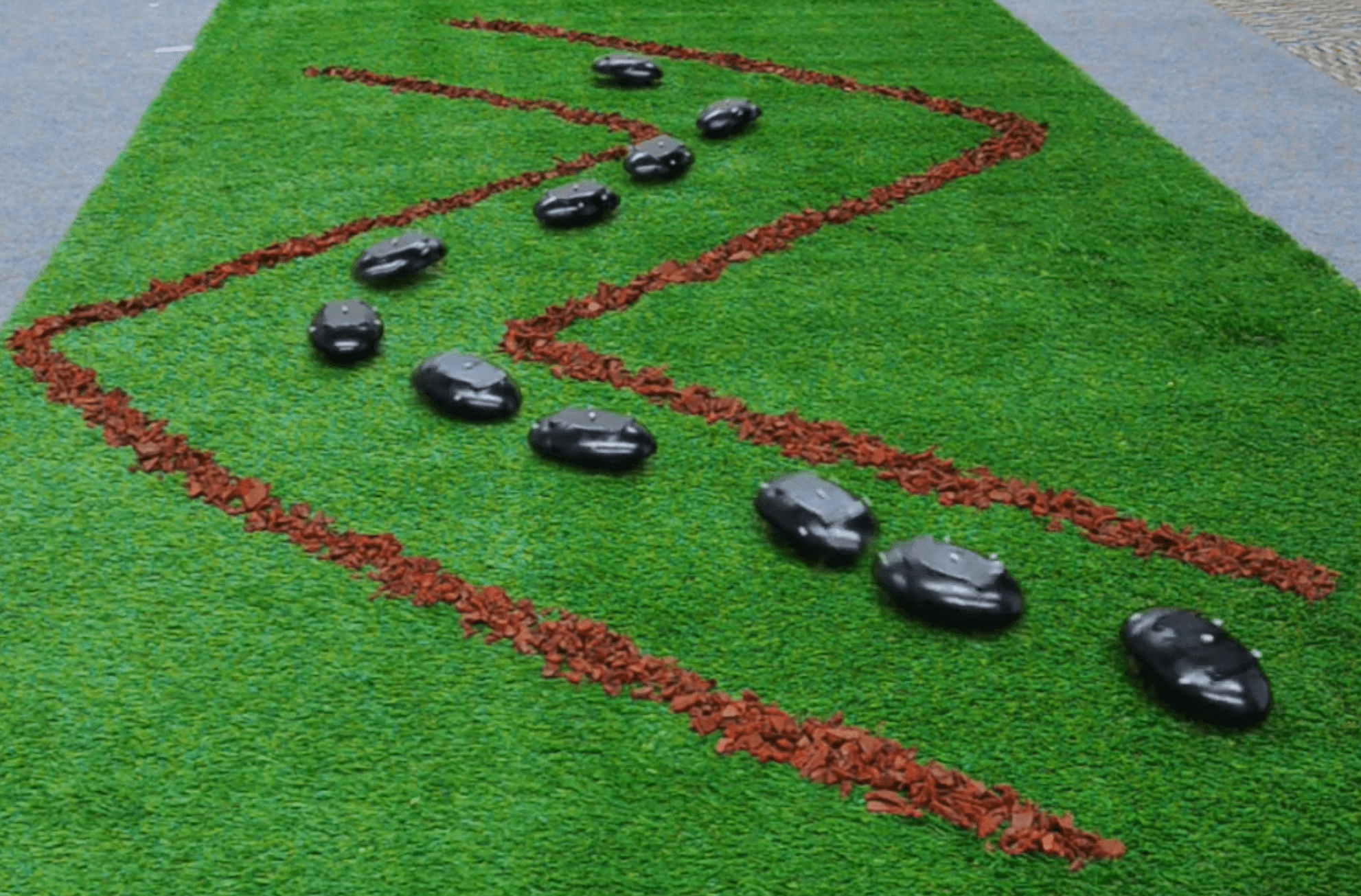}
    \caption{Execution of our model-based learning method, using an image-conditioned dynamics model, on various desired paths on the four terrains (styrofoam, gravel, carpet, and turf) we consider. Note that the path boundaries are outlined for visualization purposes only, and were not present during the experiments.}
    \label{fig:overlay}
\end{figure*}

The goal of our experimental evaluation is to study how well our model-based learning algorithm can control a real-world VelociRoACH to follow user-defined paths on various surfaces.


\subsection{VelociRoACH Platform}

The VelociRoACH is a minimally actuated, small, legged, and highly dynamic palm-sized robotic platform~\cite{haldane2013animal}. Compared to wheeled/treaded robots of similar size (Fig.~\ref{fig:wheeled}), this legged system is able to successfully navigate over more complex terrains.

The VelociRoACH is constructed through a rapid manufacturing process known as smart composite microstructure (SCM) process~\cite{scm}. This process allows for the creation of lightweight linkages, enabling the rapid realization of fully functional prototypes of folded flexure-based mobile millirobots. The VelociRoACH's robot chassis can be constructed for just \$2, and this rigid structural core houses the battery, two motors, transmission, microcontroller, and all sensors. The core also provides mechanical grounding points for the kinematic linkages, which couple each of the two motors to three legs in order to reduce the number of required actuators. 

The VelociRoACH carries an ImageProc embedded circuit board\footnote{\url{https://github.com/biomimetics/imageproc_pcb}}, which includes a 40 MHz Microchip dsPIC33F microprocessor, a six axis inertial measurement unit (IMU), an 802.15.4 wireless radio (XBee), and motor control circuitry. We added a 14-bit magnetic rotary encoders to the motors on each side of the robot to monitor absolute position. Additional sensory information includes battery voltage and back-EMF signals from the motors. 

The onboard microcontroller runs a low-level 1 kHz control loop and processes communication signals from the XBee. Due to computational limits of the microprocessor, we stream data from the robot to a laptop for calculating controller commands, and then stream these commands back to the microprocessor for execution. To bypass the problem of using only on-board sensors for state estimation, we also use an OptiTrack motion capture system to stream robot pose information during experiments. The motion capture system does not provide any information about the environment terrain, so we also mounted a 3.4 gram monocular color camera onto the VelociRoACH, which communicates directly with the laptop via a radio frequency USB receiver.


\subsection{Details of Our Approach}

The learned dynamics function $\hat{f}_\theta(\bs_t, \ba_t, \bim_t)$ is the neural network depicted in Fig.~\ref{fig:terrain-conditional}. For all experiments and results reported below, we use only 17 minutes (10,000 datapoints) of data from each terrain to train the dynamics model: This consists of 200 rollouts, each containing 50 data points that are collected at 10 Hz. We train each dynamics model for 50 epochs, using the Adam optimizer~\cite{Kingma2014_ICLR} with learning rate 0.001 and batchsize 1000.

Relevant parameters for our model-based controller are the number of candidate action sequences sampled at each time step $N=500$, the amount of time represented by one time step $\dt=0.1$ sec, the horizon $H=4$,
and parameters $f_p=50$, $f_f=10$, and $f_h=5$ for the perpendicular, forward, and heading components of the cost function from Eqn.~\ref{eqn:cost}. To simplify training and testing of the image-conditioned dynamics model, the image at the start of the rollout was used for all timesteps. The process of using the neural network dynamics model and the cost function to select the best candidate action sequence at each time step can be done in real-time, even on a laptop with no GPU, and even taking bi-directional communication delays into account.

Note that the training data is gathered entirely using random trajectories, and therefore, the paths executed by our controller at run-time differ substantially from the training data. This illustrates that our approach can be trained with off-policy data, and that the model exhibits considerable generalization. Furthermore, although the model is trained only once, we use it to accomplish a variety of tasks at run-time by simply changing the desired path in our cost function. This eliminates the need for task-specific training, which further improves overall sample efficiency. We show in Fig.~\ref{fig:overlay} some images of the VelociRoACH using our model-based learning method to execute different paths on various surfaces.


\subsection{Comparing to Differential Drive}

To provide a comparison of our model-based learning algorithm's performance, we compare to a differential drive controller, which is a common steering method used for robots with wheel or leg-like mechanisms on both sides. A differential drive control strategy controls the system's heading by specifying the left and right leg velocities based on the system's perpendicular distance to the desired path: Moving the right wheels would turn the robot to the left, and moving the left wheels would turn the robot to the right. 

In comparing our method to the differential drive controller, we tuned the differential drive controller hyperparameters in the same single environment that the model-based controller hyperparameters were tuned in. Also, all cost numbers reported below are calculated on the same cost function (Eqn.~\ref{eqn:cost}) that indicates how well the executed path aligns with the desired path, and each reported number represents an average over 10 runs.

\begin{figure}
    \centering
    \includegraphics[width=\columnwidth]{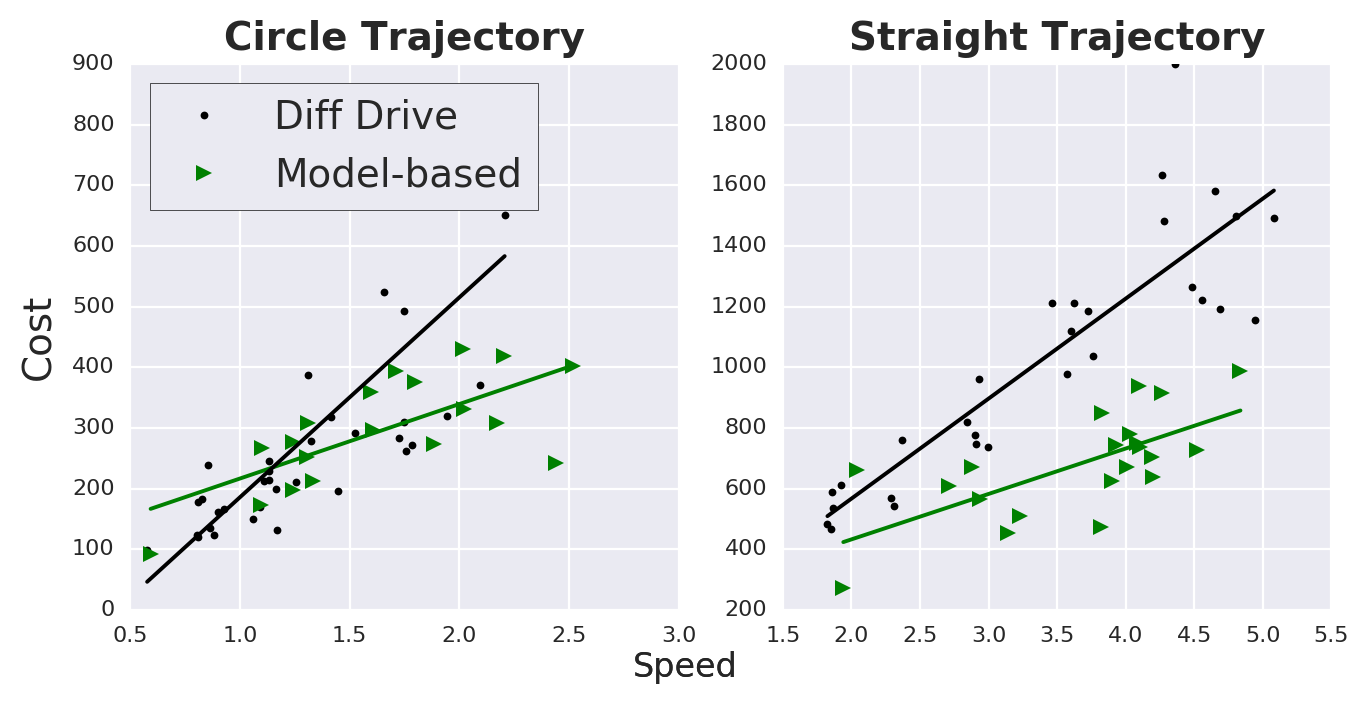}
    \caption{An analysis of cost incurred during trajectory following, as a function of the speed of the robot, shows that our model-based learning method is comparable to a differential drive control strategy at low speeds, but outperforms differential drive at high speeds.}
    \label{fig:costvsspeed}
\end{figure}

Fig.~\ref{fig:costvsspeed} illustrates, on different paths executed on carpet, that our model-based learning method and the differential drive control strategy are comparable at low speeds. However, our model-based approach outperforms the differential drive strategy at higher speeds. The performance of differential drive deteriorate as leg speeds increase, because traction decreases and causes the legs to have less control over heading. Also, at high speeds, the dynamics of the legged robot can produce significant roll oscillations, depending on the leg phasing~\cite{haldane2014roll}. Therefore, based on the timing of left and right foot contacts, the system can produce turns inconsistent with a differential drive control strategy. Fig.~\ref{fig:all_data} illustrates that for different paths across various surfaces, our model-based learning method outperforms the differential drive control strategy. Furthermore, we note that this difference in performance is most pronounced on surfaces with less traction, such as styrofoam and carpet.

\subsection{Improving Performance with More Data}

To investigate the effect of the quantity of training data, we trained three different dynamics models using different amounts of training data on carpet. We trained one with 50 rollouts (4 minutes), one with 200 rollouts (17 minutes), and one with 400 rollouts (32 minutes). Table~\ref{table_moredata} indicates that more training data can indeed improve task performance. This is an encouraging indication that improvement can occur over time, which is not the case for hand-engineered solutions.

\renewcommand{\arraystretch}{1.2}
\begin{table}[!h]
\centering
\begin{tabular}{|l|c|c|c|c|}
\hline & Straight & Left & Right\\
\hline \specialcell{50 rollouts} &14.4 &16.6 &29.4\\
\hline \specialcell{200 rollouts} &10.3 &13.6 &17.1\\
\hline \specialcell{400 rollouts} &10.8 &11.3 &11.5\\
\hline
\end{tabular}
\caption{Cost incurred by the VelociRoACH during the task of trajectory following on carpet. Three models were trained, each with different amounts of training data, and they show performance improvements occurring over time (with more data). Here, one rollout corresponds to 50 timesteps or 5 seconds of data.}
\label{table_moredata}
\end{table}


\subsection{Learning Environmental Information}

To verify whether our learned model encapsulates information about the environment, and to see whether or not the learned model itself has a large effect on controller performance, we conducted experiments on a carpet material and a slippery styrofoam material. Table~\ref{table_surfaces} shows that the baseline differential drive controller performs relatively poorly on both surfaces. For the model-based approach, the model trained on the carpet works well on the carpet, and the model trained on the styrofoam works well on the styrofoam. The poor performance of either model on the other surface illustrates that our learned dynamics model does in fact encode some knowledge about the surface. Also, performance diminishes when the model is trained on data gathered from both terrains, which indicates that this na\"{i}ve method for training a joint dynamics model is insufficient. 

\renewcommand{\arraystretch}{1.2}
\begin{table}[!ht]
\centering
\begin{tabular}{|l|c|c|c|}
\hline & Carpet & Styrofoam\\
\hline \specialcell{Differential Drive} &13.85 &15.45\\
\hline \specialcell{Model trained on carpet} &5.69 &18.62\\
\hline \specialcell{Model trained on styrofoam} &22.25 &8.15\\
\hline \specialcell{Model trained on both} &7.52 &15.76\\
\hline
\end{tabular}
\caption{Costs incurred by the VelociRoACH while executing a straight line path. The model-based controller has the best performance when executed on the surface that it was trained on. Additionally, a model trained on carpet fails on styrofoam (and vice versa), indicating that the model incorporates some knowledge about the environment of operation. Furthermore, a model trained jointly on data from all surfaces does not result in good performance.}
\label{table_surfaces}
\end{table}


\subsection{Image-Conditioned Dynamics Models}

We have shown so far that when trained on data gathered from a single terrain, our model-based approach is superior to a standard differential drive approach, and that our approach improves with more data. However, although we saw that the robot's dynamics depend on the environment, we would like our approach to be able to control the VelociRoACH on a variety of terrains.

A standard approach would be to train a dynamics model using data from all terrains. However, as shown above in Table~\ref{table_surfaces} as well as below in Fig.~\ref{fig:all_data}, a model that is na\"{i}vely trained on all data from multiple terrains and then tested on one of those terrains is significantly worse than a model that is trained solely on that particular terrain. The main reason that this na\"{i}ve approach does not work well is that the dynamics themselves differ greatly with terrain, and a dynamics model that takes only the robot's current state and action as inputs receives a weak and indirect signal about the robot's environment.

To have a direct signal about the environment, our image-conditioned model takes an additional input: an image taken from an onboard camera, as described in Sec.~\ref{sec:terrainCond}. We compare our image-conditioned dynamics model to various alternate approaches, including (a) training a separate dynamics model on each terrain, (b) na\"{i}vely training one joint dynamics model on all training data, with no images or labels, and (c) training one joint dynamics model using data with explicit terrain labels $\benv$ (Fig.~\ref{fig:terrain-conditional}) in the form of a one-hot vector (where the activation of a single vector element corresponds directly to operation in that terrain).

Fig.~\ref{fig:all_data} compares the performance of our image-conditioned approach to that of these alternative approaches, on the task of path following for four different paths (straight, left, right, zigzag) on four different surfaces (styrofoam, carpet, gravel, turf). The na\"{i}ve approach for training one joint dynamics model using an aggregation of all data performs worse than the other learning-based methods. The method of having a separate dynamics model for each terrain, as well as the method of training one joint dynamics model using one-hot vectors as terrain labels, both perform well on all terrains. However, both of these methods require human supervision to label the training data and to specify which terrain the robot is on at test time. In contrast, our image-conditioned approach performs just as well as the separate and one-hot models, but does not require any additional supervision beyond an onboard monocular camera. Finally, our image-conditioned approach also substantially outperforms the differential drive baseline on all terrains.

\begin{figure}[t]
    \centering
    \includegraphics[width=0.95\columnwidth]{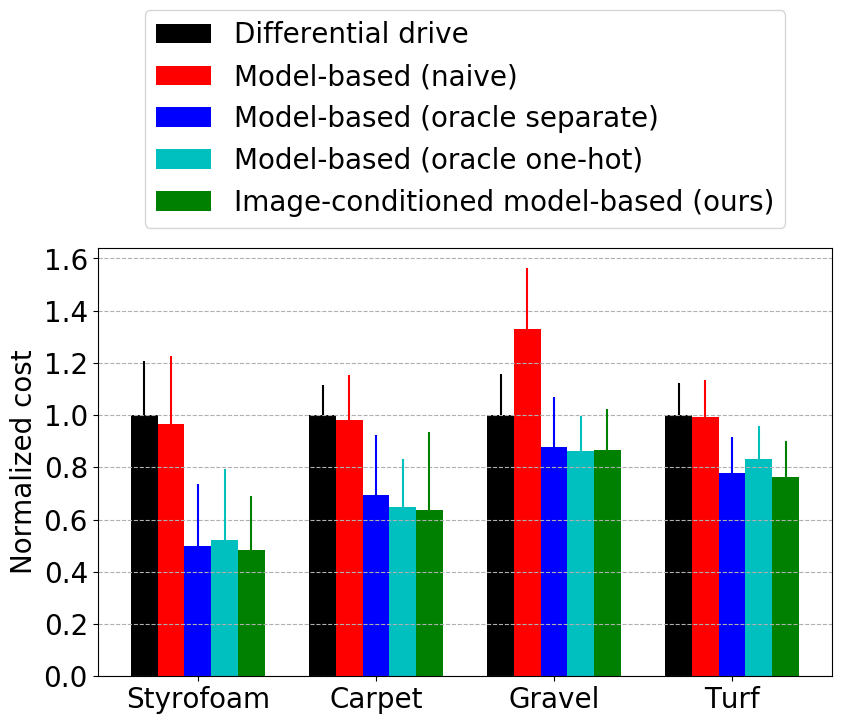}
    \caption{Comparison of our image-conditioned model-based approach to alternate methods. Each method was evaluated on four different terrains: styrofoam, carpet, gravel, and turf. On each terrain, four different paths (straight, left, right, and zigzag) were evaluated 10 times each. The methods that we compare to include: a hand-engineered differential drive controller, a joint dynamics model that is na\"{i}vely trained on all data from all terrains, an ``oracle" approach that uses a separate dynamics model on each terrain, and another ``oracle" approach where the joint dynamics model is trained using data containing an extra one-hot vector input indicating the terrain label of each data point. Our method outperforms the differential drive method and the na\"{i}ve model-based controller, while performing similarly to the oracle baselines without needing any explicit labels.}
    \label{fig:all_data}
\end{figure}

\section{Discussion}
\label{sec:discussion}
We presented a sample-efficient model-based learning algorithm using image-conditioned neural network dynamics models that enables accurate locomotion of a low-cost, under-actuated, legged, and highly dynamic VelociRoACH robot in a variety of environments. Using only 17 minutes of real-world data for each terrain, our method outperformed a commonly used differential drive control strategy, showed improvement with more data, and was able to use features from camera images in order to execute successful locomotion on various terrains.

Although our model-based controller can be repurposed at run-time to execute different paths, a drawback of this controller is the amount of computation involved at each step. We overcame the limitations of our embedded processor by streaming information to and from an external computer. However, performing all computations on-board would reduce delays, increase robustness to communication issues, and make this system more suitable for real-world tasks. One option could be to use the learned dynamics model to simulate rollouts of training data, which could then be used to train a policy without the need for more real-world data collection. However, this would require further algorithmic development, because the current dynamics model diverges after a few time steps, which precludes its direct applicability to traditional reinforcement learning algorithms that require longer rollouts for policy training. 

Another direction for future work includes developing an algorithm for online adaptation of the learned model. This would improve performance because over time, the roach suffers from deterioration of the chassis, motor strength, and leg characteristics. Furthermore, online adaptation would allow the robot to succeed at test tasks further away from the training distribution, allowing for adaptation to both new tasks as well as to unexpected environmental perturbations. Additionally, removing the dependence on a motion capture system is compelling, particularly when aiming for real-world application.

Another interesting line of future work includes improving the MPC controller. Our current approach samples random actions at each time step and uses the predictions from the dynamics model to select the action sequence with the lowest associated cost. However, sampling based approaches are intractable for systems with high-dimensional action spaces over long time horizons. Furthermore, a more structured search of the action space could prevent rapidly changing actions, limit the search space to more meaningful options, and also enable the discovery of gaits through imposing cyclic or other intelligible constraints.

\section{Acknowledgements}

We would like to thank Carlos Casarez for extensive in-lab training of VelociRoACH construction, constant mechanical support, general expertise with these systems, and many insightful discussions. This work is supported by the National Science Foundation under the National Robotics Initiative, Award CMMI-1427096.

\bibliographystyle{IEEEtran}
\bibliography{2018_iros_roach}


\end{document}